# Shift-Equivariant Similarity-Preserving Hypervector Representations of Sequences

Dmitri A. Rachkovskij

***Abstract*—Hyperdimensional Computing (HDC), also known as Vector-Symbolic Architectures (VSA), is a promising framework for the development of cognitive architectures and artificial intelligence systems, as well as for technical applications and emerging neuromorphic and nanoscale hardware. HDC/VSA operate with hypervectors, i.e., distributed vector representations of large fixed dimension (usually > 1000). One of the key ingredients of HDC/VSA are the methods for encoding data of various types (from numeric scalars and vectors to graphs) into hypervectors. In this paper, we propose an approach for the formation of hypervectors of sequences that provides both an equivariance with respect to the shift of sequences and preserves the similarity of sequences with identical elements at nearby positions. Our methods represent the sequence elements by compositional hypervectors and exploit permutations of hypervectors for representing the order of sequence elements. We experimentally explored the proposed representations using a diverse set of tasks with data in the form of symbolic strings. Although our approach is feature-free as it forms the hypervector of a sequence from the hypervectors of its symbols at their positions, it demonstrated the performance on a par with the methods that apply various features, such as subsequences. The proposed techniques were designed for the HDC/VSA model known as Sparse Binary Distributed Representations. However, they can be adapted to hypervectors in formats of other HDC/VSA models, as well as for representing sequences of types other than symbolic strings.**

## I. Introduction

HYPERDIMENSIONAL Computing (HDC [1]), also known as Vector-Symbolic Architectures (VSA [2]), is an approach that has been proposed to combine the advantages of distributed vector representations and symbolic structured data representations in Artificial Intelligence, Machine Learning, and Pattern Recognition problems. HDC/VSA have demonstrated potential in technical applications and cognitive architectures and are well-suited for implementation in the emerging stochastic, nanoscale, and neuromorphic hardware (e.g., [3]–[11] and references therein).

HDC/VSA operate with hypervectors (the term proposed in [1]), i.e., distributed vector representations of large fixed dimension. To be useful in applications (e.g., in various types of similarity search, in linear models for classification, approximation, etc.), hypervectors must be formed in such a way that they are similar for similar data. Methods for obtaining hypervectors for data of various types have been proposed, from numeric scalars and vectors to graphs [12]–[15].

A widespread data type is sequences and, in particular, symbol strings. Sequences and strings are used to represent genome and proteome, signals, textual data, computer logs, etc. Applications that benefit from sequential data representation include bioinformatics, text retrieval and near-duplicate detection, spam identification, virus and intrusion detection, spell checking, signal processing, speech and handwriting recognition, error correction, and many others (e.g., [16], [17] and references therein).

The methods of similarity search, clustering, classification (both by similarity search and by various parametric methods), etc., require an assessment of the similarity of sequences. Formation of hypervector representations that reflect the similarity of sequences opens up the possibility of using a large arsenal of methods developed specifically for vectors. These are methods of statistical pattern recognition, linear and nonlinear methods of classification and approximation, index structures for fast similarity search, selection of informative features, and others.

There are several techniques for the representation of sequences with hypervectors. However, neither of them satisfies both the requirement of equivariance with respect to the sequence shift and the requirement of preserving the similarity of sequences with identical elements at nearby positions (see section III). In this paper, we propose an approach for hypervector representation of symbol sequences that satisfies these two requirements. Our methods are based on the use of hypervector permutations to represent the order of sequence elements and were developed for the HDC/VSA model of Sparse Binary Distribution Representations [14], [15], [18] (SBDR). However, the proposed approach can be adapted for hypervector formats of other HDC/VSA models, as well as for representing sequences of other types.

Manuscript first submission August 19, 2021.
Manuscript second submission September 16, 2021
Minor revision December 31, 2021.
This work was supported in part by the by the National Academy of Sciences of Ukraine grants 0120U000122, 0121U000016, 0117U002286 and the Ministry of Education and Science of Ukraine.

D. A. Rachkovskij is with the International Research and Training Center for Information Technologies and Systems, Department of Neural Information Processing Technologies, 03680 Kiev, Ukraine (e-mail: dar@infrm.kiev.ua).

The main contributions of this paper are as follows:

1. Permutation-based hypervector representation of sequences that is shift-equivariant and preserves the similarity of sequences with the same elements at nearby positions.
2. Measures of hypervector similarity of sequences.
3. Measures of symbolic similarity of sequences that approximate the proposed hypervector similarity measures.
4. Experimental study of the proposed hypervector representations of sequences and similarity measures in diverse tasks.

## II. Background and Basic Notions

### A. Hyperdimensional computing

In various HDC/VSA models, hypervector (HV) components have a different format. For example, they can be real numbers from the Gaussian distribution (the HRR model [19]) or binary values from {0,1} (the BSC model [20] and SBDR). Data HVs are formed from the hypervectors of the data elements, usually without changing the HV dimension. E.g., for elements-symbols their hypervectors are i.i.d. randomly generated vectors of high dimension $D$, commonly $D > 1000$. Such random HVs are considered dissimilar. The similarity of HVs is usually measured based on their (normalized) dot product. In a particular task, the same data object is represented by its fixed HV.

A set of data objects (e.g., a set of symbols) is represented by the "superposition" of their HVs, e.g., by component-wise addition for real-valued HVs, or addition followed by thresholding for binary HVs. Superposition does not preserve information about the order or grouping of the objects. The HV of superposition is similar to the hypervectors that are superimposed.

To represent a sequence of data objects, their HVs are modified in a special way. For instance, for a hypervector representation of a symbol at some position, the HV of that position ("role") is "bound" to the HV of the symbol ("filler"). Binding can be performed, e.g., by component-wise conjunction (in SBDR) or by XOR (in BSC) for binary HVs, or by cyclic convolution for real-valued HVs (in HRR). This type of binding is called "multiplicative" binding. In another, "permutative" binding type, a role is represented not by a HV, but by a (random) permutation of dimension $D$, fixed for the particular role, which is applied to the filler HV. A hypervector resulting from binding contains information about the HVs from which it is formed – e.g., about the role and the filler. Binding operation distributes over superposition operation.

Most of the binding operations produce dissimilar HVs for the case when dissimilar filler HVs are bound with the same role, or when the same filler HV is bound with dissimilar roles. "Dissimilar" means that the similarity value is of the order of that for random HVs. For the bound HVs to be similar, both the HVs of the roles as well as the HVs of the fillers should be similar.

The obtained bound HVs are then superimposed. The resulting HV contains information about the bound HVs superimposed, and the bound HVs, in their turn, contain information about their respective constituents. For example, the HV of a symbol string is formed as a superposition of HVs that result from binding HVs of its symbols and HVs of their positions in the string. Known schemes for hypervector representation of strings are given in section III, and those proposed in this work appear in section IV.

This paper uses the HDC/VSA model of SBDR. In SBDR, binary hypervectors are used with a small number of $M << D$ (randomly placed) 1-components, the rest of the components are 0s. Superposition is performed by component-wise disjunction. Though the multiplicative binding procedures exist for SBDR [21], in this paper we only use permutative binding.

### B. Symbol sequences and their similarity measures

We will consider sequences of symbols from a finite alphabet. The symbol (sequence element) $x$ at position $i$ is denoted as $x_i$. E.g., $a_0$ denotes $a$ at the beginning of the string (at the initial position), $a_{-1}$ is the same symbol shifted one position left, $b_3$ is the symbol $b$ shifted 3 positions right, and so on. If a symbol is specified without an index, it is at the initial position: $x \equiv x_0$.

We denote by $x_i y_j \ldots z_k$ the sequence of symbols $x, y, \ldots, z$ at positions, respectively, $i, j, \ldots, k$, e.g., $b_3 c_1 a_4 a_{-3}$. A symbol string (symbols at consecutive positions) is denoted as $x_i y_{i+1} \ldots z_{i+k} \equiv xy \ldots z_i$, e.g., $c_1 b_2 c_3 a_4 \equiv cbca_1 \equiv (cbca)_1$. A string without an index is at its initial position, e.g., $cbca \equiv cbca_0 \equiv c_0 b_1 c_2 a_3$.

Various similarity measures are used for strings [17]. The Hamming distance $\text{dist}_{\text{Ham}}$ is equal to the number of non-matching symbols at the same positions ($\text{sim}_{\text{Ham}}$ is defined as the number of matching symbols). Hamming measures capture the intuitive idea of string similarity: the similarity of a string to itself is greater than to other strings, and the more is the number of mismatchings, the less similar strings are, e.g., $\text{sim}_{\text{Ham}}(cbca_0, cbca_0) > \text{sim}_{\text{Ham}}(cbca_0, cbcb_0) > \text{sim}_{\text{Ham}}(cbca_0, cbab_0)$.

Hamming similarity can be extended to strings of different lengths by augmenting a shorter string with special symbols. One can also compare strings at different positions, for example, $cbca_0$ and $cbca_1$, by representing them as $c_1 b_2 c_3 a_4 \$_5$ and $\$_1 c_2 b_3 c_4 a_5$, where \$ is a special symbol that does not belong to the alphabet of string symbols. The last example, however, breaks the intuition of string similarity, since $\text{sim}_{\text{Ham}}(c_1 b_2 c_3 a_4 \$_5, \$_1 c_2 b_3 c_4 a_5) = 0$, however, these strings seem to be similar to us. This problem is solved by the shift distance [22], defined as the minimum Hamming distance between one string and some cyclic shift of the other string.

An alternative approach to string comparison is the Levenshtein distance $\text{dist}_{\text{Lev}}$ defined as the minimum number of edit operations required to change one string into the other [23]. For $\text{dist}_{\text{Lev}}$, edit operations are symbol insertion, deletion, and substitution. The complexity of calculating $\text{dist}_{\text{Lev}}$ (by dynamic programming) is quadratic of the string length. $\text{dist}_{\text{Lev}}$ is widely used in practice, so methods of speeding up its estimation and similarity search are a direction of intensive research [16], [17], [24].

Alignment-free sequence comparison methods [25] do not use dynamic programming to "align" the whole strings (i.e., to find a match between all symbols in two strings) and their computational complexity is sub-quadratic. The methods are based on $n$-gram frequencies, the length of common substrings, the alignment of substrings, the use of words with some symbol gaps, etc.

### C. Equivariance of hypervectors with respect to sequence shift

Let $x$ be an object (input), $F$ be a function performing a representation, $F(x)$ be the result of $x$ representation. $F$ is equivariant with respect to transformations $T$, $S$ if [26]: $F(S(x)) = T(F(x))$. Transformations $T$, $S$ can be different. If $T$ is the identity transformation, $F$ is invariant with respect to $S$.

We consider the hypervector representation of sequences. Let us represent the sequence $x$ as a HV by applying some function (algorithm) $F(x)$. Then shift $x$ to another position, denote this transformation by $S(x)$. The hypervector of the shifted sequence is obtained as $F(S(x))$. The representation function $F$ equivariant with respect to $S(x)$ must ensure $F(S(x)) = T(F(x))$, where $T$ is some transformation of the hypervector $F(x)$. In other words, the hypervector of the shifted sequence can be obtained not only by transforming this sequence into a hypervector, but also by some transformation of the HV of the unshifted sequence.

Hypervectors corresponding to symbols/sequences will be denoted by the corresponding bold letters. E.g., $F(a_0) = \mathbf{a}_0$, $F(cbca_0) = \mathbf{cbca}_0$. $F(cbca_s) = \mathbf{cbca}_s$. We denote the shift of symbols by $s$ positions by $S_s$: $S_1(a_0) = a_1$, $S_{-1}(abc) = S_{-1}(abc_0) = abc_{-1} \equiv a_{-1}b_0c_1$. Let $T_s$ denote the hypervector transform corresponding to $S_s$. To ensure equivariance, the following must be true: $F(S_s(x)) = T_s(F(x)) = \mathbf{x}_s$ ($x$ is a symbol or sequence). E.g., for specific symbols or strings: $F(S_1(a_0)) = \mathbf{a}_1 = T_1(F(a_0))$, $F(S_2(abc_0)) = T_2(F(abc_0)) = \mathbf{abc}_2$, $F(S_{-2}(abc_4)) = T_{-2}(F(abc_4)) = \mathbf{abc}_2$, etc. In section III-C, hypervector representations of sequences are shown that are shift-equivariant and use a permutation as $T$.

## III. Related Work

As mentioned in section II-A, in the HDC/VSA-based methods for representing sequences, each element of a sequence is associated with a hypervector. For symbol strings, symbols are considered dissimilar and so they are assigned randomly generated (and thereafter fixed) hypervectors in the format of the HDC/VSA model being used. To represent sequence elements at their positions, element HVs are modified in various ways. In [27], the following modifications were identified: multiplicative binding with the position HV, multiplicative binding with the HVs of other (e.g., context) elements, and binding the element HV with its position by permutation. $N$-gram representations are also used. Below, we review some of these approaches in more detail. Let us show that the hypervectors formed by these approaches either do not preserve the similarity of sequences with identical elements at nearby positions or are not shift-equivariant.

### A. Multiplicative binding with position

In [28], it was proposed to bind the hypervectors of symbols with the hypervectors of their positions by a multiplicative binding operation (section II-A). Thus, the HV of the sequence $x_iy_j \ldots z_k$ is formed as $\mathbf{x}_i\mathbf{y}_j \ldots \mathbf{z}_k = F(x_iy_j \ldots z_k) = \mathbf{x} \otimes \mathbf{pos}_i \oplus \mathbf{y} \otimes \mathbf{pos}_j \oplus \ldots \oplus \mathbf{z} \otimes \mathbf{pos}_k$, where $\mathbf{pos}_k$ is the HV of the $kth$ position, $\otimes$ is the binding operation, $\oplus$ is the superposition operation. For example, for the string $abc$, the hypervector is formed as $\mathbf{abc} = \mathbf{a} \otimes \mathbf{pos}_0 \oplus \mathbf{b} \otimes \mathbf{pos}_1 \oplus \mathbf{c} \otimes \mathbf{pos}_2$. Such a representation was also considered in [19] and was applied, e.g., in [29], [30]. I.i.d. random hypervectors for positions were used. This representation does not preserve the similarity of symbol hypervectors at nearby positions and is not shift-equivariance.

The following approach allows obtaining shift-equivariance. Some multiplicative binding operations allow recursive binding of a hypervector to itself [19], e.g.: $\mathbf{pos}^j = \mathbf{pos} \otimes \ldots_{(j \text{ times})} \ldots \otimes \mathbf{pos}$. The representation of the sequence in the form $\mathbf{x}_i\mathbf{y}_j \ldots \mathbf{z}_k = F(x_iy_j \ldots z_k) = \mathbf{x} \otimes \mathbf{pos}^i \oplus \mathbf{y} \otimes \mathbf{pos}^j \oplus \ldots \oplus \mathbf{z} \otimes \mathbf{pos}^k$ allows obtaining the hypervector of the shifted sequence as (using the distributivity of the binding operation over the superposition):

$$F(S_s(x_iy_j \ldots z_k)) = F(x_{i+s}y_{j+s} \ldots z_{k+s}) = \\ \mathbf{pos}^{i+s} \otimes \mathbf{x}_0 \oplus \mathbf{pos}^{j+s} \otimes \mathbf{y}_0 \oplus \ldots \oplus \mathbf{pos}^{k+s} \otimes \mathbf{z}_0 = \\ \mathbf{pos}^s \otimes (\mathbf{pos}^i \otimes \mathbf{x} \oplus \mathbf{pos}^j \otimes \mathbf{y} \oplus \ldots \oplus \mathbf{pos}^k \otimes \mathbf{z}) = T_s(F(x_iy_j \ldots z_k)). \quad (1)$$

Thus, such a hypervector of a string is equivariant with respect to the string shift for $T_s = \mathbf{pos}^s \otimes$. However, this HV does not preserve the similarity of a symbol at nearby positions, since the position hypervectors are not similar and therefore $\mathbf{pos}^i \otimes \mathbf{x}$ is not similar to $\mathbf{pos}^j \otimes \mathbf{x}$ for $i \neq j$. The MBAT approach [31] has similar properties, however, the position binding is performed by multiplying by a random position matrix.

### B. Multiplicative binding with correlated position hypervectors

As mentioned in [27], [32], if the position hypervectors are similar (correlated) for nearby positions, the hypervector representation preserves the similarity of the symbol at different nearby positions. The binding with correlated roles represented by correlated random matrices was proposed in [33]. We are not aware of transformations that ensure shift-equivariance of such string hypervectors.

Based on the ideas of [19], in [34]–[37] an approach using multiplicative binding is considered. It represents a coordinate value by converting a random hypervector into a complex one using FFT and raises the result component-wise to the fractional power corresponding to the coordinate value. The HV similarity decreases from 1 to 0 when the coordinate increases from 0 to 1. It could be adapted to the representation of strings by associating positions with small coordinate changes and ensures equivariance (mathematically, at least). However, it works with real-valued hypervectors and does not apply to binary hypervectors, and requires expensive forward and inverse FFT.

### C. Permutative binding with position

Using permutations of hypervector components to represent the order of sequence elements has been proposed in [38], [1]. The hypervector of the sequence $x_iy_j \ldots z_k$ is formed as $\mathbf{x}_i\mathbf{y}_j \ldots \mathbf{z}_k = F(x_iy_j \ldots z_k) = \mathrm{perm}_i(\mathbf{x}_0) \oplus \mathrm{perm}_j(\mathbf{y}_0) \oplus \ldots \oplus \mathrm{perm}_k(\mathbf{z}_0)$, where $\mathrm{perm}_k$ is the permutation corresponding to the $k$-th position. Similar ideas were considered in [19], [14].

Let $\mathrm{perm}_k = \mathrm{perm}^k$, where $\mathrm{perm}^k(\mathbf{x}) = \mathrm{perm}\ (\mathrm{perm}(\mathrm{perm}\ldots_k$

$_{\text{times}}$ ... perm(**x**)...)) is the sequential application of $k$ identical permutations. Here perm is usually a random permutation and $\text{perm}^0(\mathbf{x}) = \mathbf{x}$. For $k < 0$, $\text{perm}^{-|k|}(\mathbf{x})$ denotes the $k$ permutations inverse to perm. This hypervector representation of a sequence is equivariant with respect to the sequence shift:

$$F(S_s(x_i y_j \ldots z_k)) = F(x_{i+s} y_{j+s} \ldots z_{k+s}) = \text{perm}^{i+s}(\mathbf{x}_0) \oplus \text{perm}^{j+s}(\mathbf{y}_0) \oplus \ldots \oplus \text{perm}^{k+s}(\mathbf{z}_0) = \text{perm}^{s}(\text{perm}^{i}(\mathbf{x}) \oplus \text{perm}^{j}(\mathbf{y}) \oplus \ldots \oplus \text{perm}^{k}(\mathbf{z})) = T_s(F(x_i y_j \ldots z_k)). \quad (2)$$

However, such a representation does not preserve the similarity of the same symbols at nearby positions, since permutation does not preserve the similarity of the permuted hypervector with the original one.

In [39], the representation of a word was formed from the hypervectors of its letters cyclically shifted by the number of positions corresponding to the letter position in the word. In addition, to preserve the similarity with the words containing the same letters in a different order, the original hypervectors of letters were superimposed into the final hypervector of the word. However, shifting this hypervector would give the hypervector different from that obtained by superimposing the initial letter hypervectors with the hypervectors of the shifted word letters at their positions.

### D. Binding by partial permutations

To preserve the similarity of hypervectors when using permutations, [40] proposed to use partial (correlated) permutations. Let us apply this approach to symbol sequences. Symbols are represented by random sparse binary HVs **x**. The HV of a symbol at position $i$ is formed as follows. Let $R$ ("similarity radius") be an integer. The **x** is permuted $\lfloor i/R \rfloor$ times. Then we additionally permute a part of 1-components of the resulting HV, the part being equal to $i/R - \lfloor i/R \rfloor$. The rest of the 1-components coincide with the 1-components of the HV at the position $R\lfloor i/R \rfloor$. HVs of all string symbols obtained in this way are superimposed (by component-wise disjunction). For the HV of a symbol at positions $i,j$, this method approximates the linearly decreasing similarity characteristic $1 - |i - j| / R$ for $|i - j|/ < R$. For $|i - j|/ \geq R$, the similarity is close to 0 (corresponds to the similarity of random hypervectors). Such a decreasing similarity is also observed for the HV of a sequence.

When forming the sequence HV, since all the sequence symbols are at different positions, their HVs are permuted in different ways. Therefore, when shifting the string, the HV of each symbol must be permuted differently, taking into account the current position of the symbol. However, we cannot do this, since we have access only to the holistic hypervector of the whole string. Thus, equivariance is not ensured. We are forced to calculate the new positions of symbols in the shifted string $x_{i+s} y_{j+s} \ldots z_{k+s}$ and re-form the hypervector of the sequence at a new position from the scratch: $F(S_s(x_i y_j \ldots z_k)) = F(x_{i+s} y_{j+s} \ldots z_{k+s}) = \mathbf{x}_{i+s} \vee \mathbf{y}_{j+s} \vee \ldots \vee \mathbf{z}_{k+s}$. Shift-equivariance is also absent in [41], where partial permutations of dense hypervectors are used.

## IV. Method

To preserve both the equivariance of hypervector representations of sequences with respect to the shift and the similarity of the sequence hypervectors having the same symbols at nearby positions, we propose to form the HVs of symbols as compositional HVs of a specific structure, using random permutation and superposition. We use the SBDR model (section II-A).

### A. Hypervector representation of symbols

To represent the symbol $a$, we will form its hypervector **a** as follows. Let's generate a random ("atomic") HV $\mathbf{e}_{a\,0}$. Let's form other atomic HVs as: $\mathbf{e}_{a\,i} = \text{perm}(\mathbf{e}_{a\,i-1}) = \text{perm}^i(\mathbf{e}_{a\,0})$. Obtain the hypervector of the symbol $a$ at position $i$ (that is, $\mathbf{a}_i = F(a_i)$ for a given value of $R$) as $\mathbf{a}_i = \mathbf{e}_{a\,i} \vee \mathbf{e}_{a\,i+1} \vee \ldots \vee \mathbf{e}_{a\,i+R-1}$.

*1) Equivariance:* For such a hypervector representation, the equivariance with respect to the symbol shift holds if an appropriate permutation is used as the hypervector transformation $T$. Indeed,

$$T_j(F(a_i)) = \text{perm}^j(\mathbf{a}_i) = \text{perm}^j(\mathbf{e}_{a\,i} \vee \mathbf{e}_{a\,i+1} \vee \ldots \vee \mathbf{e}_{a\,i+R-1}) = \text{perm}^j(\mathbf{e}_{a\,i}) \vee \text{perm}^j(\mathbf{e}_{a\,i+1}) \vee \ldots \vee \text{perm}^j(\mathbf{e}_{a\,i+R-1}) = \mathbf{e}_{a\,i+j} \vee \mathbf{e}_{a\,i+j+1} \vee \ldots \vee \mathbf{e}_{a\,i+j+R-1} = \mathbf{a}_{i+j} = F(a_{i+j}) = F(S_j(a_i)). \quad (3)$$

*2) Similarity:* Let us consider hypervectors $\mathbf{a}_i = \mathbf{e}_{a\,i} \vee \mathbf{e}_{a\,i+1} \vee \ldots \vee \mathbf{e}_{a\,i+R-1}$ and $\mathbf{a}_{i+j} = \mathbf{e}_{a\,i+j} \vee \mathbf{e}_{a\,i+j+1} \vee \ldots \vee \mathbf{e}_{a\,i+j+R-1}$. For $|j| < R$, $\mathbf{a}_i$ and $\mathbf{a}_{i+j}$ have $R - |j|$ coinciding atomic HVs. E.g., for $j > 0$ these are atomic HVs with the indices from $i + j$ to $i + R - 1$ (the last atomic HVs from $\mathbf{a}_i$ and the first atomic HVs from $\mathbf{a}_{i+j}$; for $j < 0$, the opposite is true). For $|j| \geq R - 1$, $\mathbf{a}_i$ and $\mathbf{a}_{i+j}$ have no coinciding atomic HVs.

For atomic hypervectors **e** with the number of 1-components $|\mathbf{e}| = m \ll D$, their intersection is small with high probability. For the case without the intersection of atomic HVs, the similarity of symbol hypervectors $\mathbf{a}_i$ and $\mathbf{a}_{i+j}$ at different positions inside $R$ is $m(R - |j|)$ (in terms of the number of coinciding 1-components). For the case with the intersection of atomic HVs, the similarity of $\mathbf{a}_i$ and $\mathbf{a}_{i+j}$ inside $R$ may somewhat vary around this value, and there may be similarities between $\mathbf{a}_i$ and $\mathbf{a}_{i+j}$ outside $R$.

### B. Hypervector representation and similarity of sequences

Hypervectors of various symbols at their positions are formed by the method of section IV-A from their randomly generated atomic HVs, using the same permutation. Generally, for random permutation, some intersection of the HVs of symbols $x_i$ и $y_j$ is possible for any $x \neq y$ and for any $i, j$. A symbol sequence HV is formed from the symbol HVs using permutation and superposition operation: $\mathbf{x}_i \mathbf{y}_j \ldots \mathbf{z}_k = F(x_i y_j \ldots z_k) = \text{perm}^i(\mathbf{x}_0) \vee \ldots \vee \text{perm}^k(\mathbf{z}_0)$.

The properties of equivariance and preservation of similarity for hypervectors of symbol sequences can be obtained in the same manner as in sections IV-A and III-C. This is achieved due to the distributive property of vector permutation over the superposition operation (the permutation distributivity is also preserved over any component-wise operation on vectors [1]).

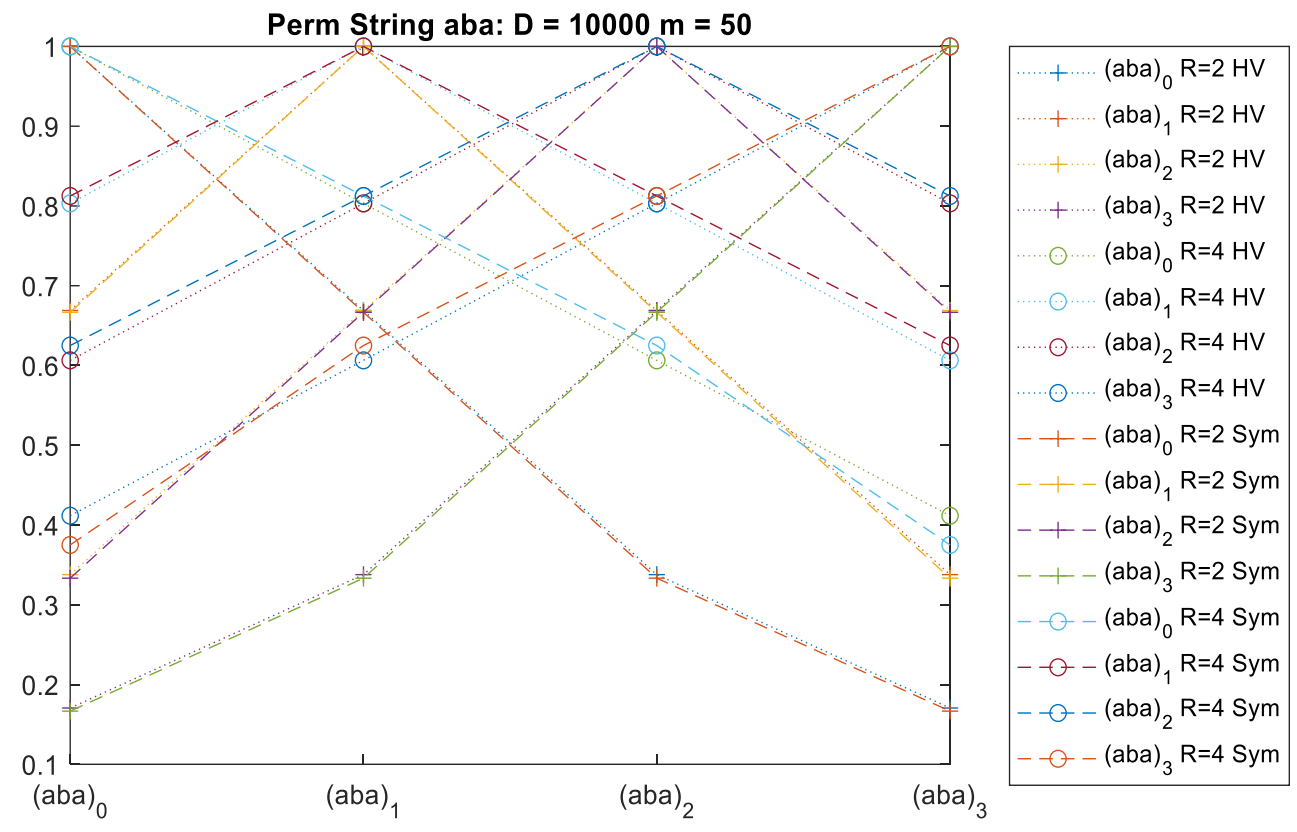

Fig. 1. The hypervector (HV) and the symbolic (Sym) similarity of *aba*.

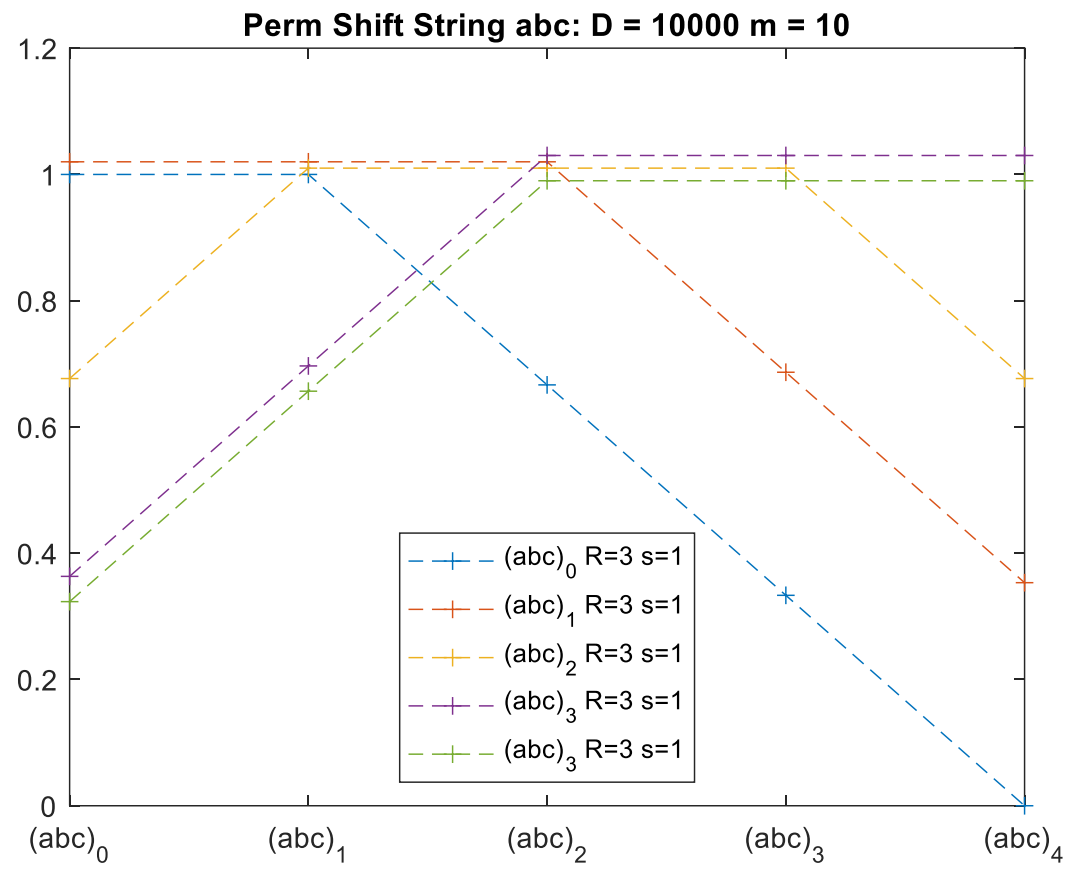

Fig. 2. The values of $\mathrm{sim}_{\mathrm{HV,cos}}$ between the hypervector representation of the string *abc* at different positions, taking into account the shift. $R = 3$, $s = 1$.

*1) Hypervector similarity of strings without shift*: The hypervector similarity is calculated using usual similarity measures of binary vectors. The normalized similarities with the values in [0,1] are given, e.g., by the following measures.

The cosine similarity: $\mathrm{sim}_{\cos} = |\mathbf{a}\wedge\mathbf{b}| / \mathrm{sqrt}(|\mathbf{a}||\mathbf{b}|) \equiv \langle\mathbf{a},\mathbf{b}\rangle / \mathrm{sqrt}(\langle\mathbf{a},\mathbf{a}\rangle\langle\mathbf{b},\mathbf{b}\rangle)$, where $|\mathbf{x}| \equiv \langle\mathbf{x},\mathbf{x}\rangle$ is the number of 1-components in $\mathbf{x}$, $\langle\cdot,\cdot\rangle$ is the dot product. Jaccard: $\mathrm{sim}_{\mathrm{Jac}} = |\mathbf{a}\wedge\mathbf{b}| / |\mathbf{a}\vee\mathbf{b}| = |\mathbf{a}\wedge\mathbf{b}| / (|\mathbf{a}|+|\mathbf{b}|-|\mathbf{a}\wedge\mathbf{b}|)$. Simpson: $\mathrm{sim}_{\mathrm{Simp}} = |\mathbf{a}\wedge\mathbf{b}| / \min(|\mathbf{a}|,|\mathbf{b}|)$.

Let us denote by $\mathrm{sim}_{\mathrm{HV},R,\mathrm{type}}(a,b)$ the measure of hypervector similarity of symbol sequences $a,b$. HVs are obtained by the method proposed above for a given $R$ value. The "type" stands for, e.g., cos, Jac, Simp, etc. This similarity measure is alignment-free, see section II-B. Examples of hypervector similarity characteristics for a string at different positions are shown in Fig. 1.

*2) Hypervector similarity of strings with the shift*: Strings might have identical substrings outside $R$. E.g., for $dddabc_0$ and $abc_0$, the value of $\mathrm{sim}_{\mathrm{HV,R,type}}$ is close to zero for $R \geq 3$. However, if $abc_0$ is shifted to $abc_3$, the string $\mathrm{abc}_3$ will match the substring of $dddabc_0$. Let us take into account such cases by calculating the similarity as the maximum value of $\mathrm{sim}_{\mathrm{HV,R,type}}$ for various shifts of one of the sequences:

$$\mathrm{sim}_{\mathrm{HV},R,s,\mathrm{type}}(a,b) \equiv \mathrm{sim}_{\mathrm{HV},R,s,\mathrm{type}}(F(a),F(b)) \equiv \mathrm{sim}_{\mathrm{HV},R,s,\mathrm{type}}(\mathbf{a},\mathbf{b}) = \max_s \mathrm{sim}_{\mathrm{HV},R,\mathrm{type}}(S_s(a),b). \quad (4)$$

Unless stated otherwise, we assume that the numeric value $s$ specifies the set of shifts from $-s$ to $s$ in steps of 1. E.g., if $s = 1$ then $\mathrm{sim}_{\mathrm{HV},R,s,\mathrm{type}}(a,b)$ is the max value of $\mathrm{sim}_{\mathrm{HV},R,\mathrm{type}}(S_s(a),b)$ obtained with shifts $\{-1,0,1\}$ of the sequence $a$, see Fig. 2. Equivariance permits obtaining the HVs of shifted sequences by permuting the sequence HV obtained for a single position. For brevity, if the values of $R$, $s$, type are clear, we denote our hypervector similarity measures as $\mathrm{sim}_{\mathrm{HV}}$.

## C. *A symbolic similarity measure for symbol sequences*

Let us introduce a symbolic similarity measure for symbol sequences that is analogous to the proposed $\mathrm{sim}_{\mathrm{HV}}$ but does not use the transformation of strings into hypervectors. We denote: symbol sequences as $a$, $b$; an element of sequence $x$ at the position $i$ as $x_i$; the similarity radius as $R \subset \mathbb{Z}_{\geq 0}$ (a fixed non-negative integer); $\delta_{ij} = |\,i - j\,|$;

$$\Delta_{i,R} = \Sigma_j \, 1 - \delta_{ij}/R \text{ if } a_i \in b \wedge \delta_{ij} \leq R \,|\, j\!: b_j = a_i; \quad \Delta_{i,R} = 0 \text{ otherwise} \quad (5)$$

Then the measure of string similarity, which we call "symbolic overlap" SymOv, is given by

$$\mathrm{sim}_{\mathrm{SymOv},R}(a,b) = \Sigma_i \, \Delta_{i,R}. \quad (6)$$

This SymOv similarity is analogous to $|\mathbf{a}\wedge\mathbf{b}|$ for hypervectors of strings $a$, $b$. To obtain normalized similarities with the values in [0,1] (analogous to $\mathrm{sim}_{\mathrm{HV}}$ from section IV-B-1), we define the SymOv-norm of a symbol sequence $x$ as $|x|_R = \mathrm{sim}_{\mathrm{SymOv},R}(x,x)$. Then different types of normalized similarities $\mathrm{sim}_{\mathrm{Sym},R,\mathrm{type}}(a,b)$ are defined analogously to $\mathrm{sim}_{\mathrm{HV},R,\mathrm{type}}(a,b)$.

Taking into account shifts, we obtain:

$$\mathrm{sim}_{\mathrm{Sym},R,s,\mathrm{type}}(a,b) = \max_s \mathrm{sim}_{\mathrm{Sym},R,\mathrm{type}}(S_s(a), b). \quad (7)$$

The values of these similarities would coincide with the (expected) values of hypervector measures, provided that the symbol HVs are superimposed in the sequence HV by addition instead of component disjunction, and $|\mathbf{x}\wedge\mathbf{y}|$ is changed to $\langle\mathbf{x},\mathbf{y}\rangle$.

# V. Experiments

Experimental evaluations of the proposed approach were carried out in several diverse tasks: spellchecking (section V-A), classification of molecular biology data (section V-B), modeling the identification of visual images of words by humans (section V-C). It is important to note that the experiments were intended to demonstrate the feasibility of the proposed approach, and so there was no goal to obtain the state-of-the-art results.

## A. *Spellchecking*

For misspelled words, the spellchecker suggests one or more variants of the correct word. We used similarity search for this problem. Dictionary words and misspelled (query) words were transformed to hypervectors by the methods of section IV. A specified number of dictionary words with the HVs most similar to the HV of a query word were selected.

TABLE I
SPELLCHECKING ACCURACY ON THE ASPELL AND WIKIPEDIA TESTS

| Study | Method | Top1 % | Top3 % | Top5 % | Top10 % | Top1 % | Top3 % | Top5 % | Top10 % |
|---|---|---|---|---|---|---|---|---|---|
| | | aspell | | | | wikipedia | | | |
| [43] | Ispell | 36.0 | 47.7 | 50.3 | 51.7 | 76.0 | 82.8 | 83.2 | 83.4 |
| [43] | Aspell (normal) | 56.9 | 74.4 | 81.0 | 87.9 | 84.7 | 95.6 | 97.4 | 98.5 |
| [43] | Word 97 | 59.0 | 69.0 | 71.0 | 72.6 | 89.0 | 94.3 | 94.7 | 95.0 |
| [43] | Word 2003 | 62.8 | 74.1 | 77.2 | 78.2 | 92.6 | 96.1 | 96.5 | 96.6 |
| [43] | Deorowicz et al. | 66.3 | 79.6 | 83.6 | 85.5 | 94.1 | 98.3 | 98.9 | 99.0 |
| [44] | Mitton | 71.1 | 88.6 | 91.4 | 94.4 | 92.9 | 97.9 | 98.6 | 99.0 |
| [45] | Omelchenko HV | 58.6 | 77.7 | 82.4 | 88.9 | 80.0 | 92.8 | 95.7 | 97.5 |
| Our | Lev | 47.8 | 67.1 | 73.9 | 82.2 | 66.1 | 81.6 | 85.7 | 90.0 |
| Our | Lev/max | 54.2 | 73.1 | 78.9 | 85.9 | 70.9 | 83.3 | 86.5 | 89.7 |
| Our | Sym R=7 | 56.7 | 74.5 | 78.9 | 85.1 | 81.2 | 94.2 | 96.4 | 97.7 |
| Our | Sym R=7 s=1 | 55.9 | 74.7 | 78.5 | 84.7 | 81.1 | 94.7 | 96.1 | 97.9 |
| Our | HV R=7 m=11 | 59.0 | 76.5 | 81.7 | 87.2 | 82.8 | 93.9 | 96.4 | 98.0 |
| | std (50) | 0.467 | 0.419 | 0.408 | 0.262 | 0.279 | 0.134 | 0.125 | 0.071 |
| Our | HV R=7 m=11 | 59.4 | 76.3 | 81.1 | 86.4 | 84.39 | 94.36 | 96.61 | 97.93 |
| | s=1 std (50) | 0.522 | 0.482 | 0.406 | 0.471 | 0.234 | 0.165 | 0.110 | 0.0778 |
| Our | HV R=7 m=11 | 59.2 | 75.8 | 80.8 | 86.3 | 83.78 | 94.37 | 96.55 | 96.86 |
| | s=2 std (50) | 0.559 | 0.449 | 0.354 | 0.479 | 0.27 | 0.177 | 0.116 | 0.0802 |

TABLE II
ACCURACY ON THE SPLICE-JUNCTION GENE SEQUENCES DATASET

| Study | Method | Total % | EI % | IE % | Neither % |
|---|---|---|---|---|---|
| [46] | Hybrid KBANN | – | 92.44 | 91.53 | 95.38 |
| [47] | C4.5 decision-tree | 95.7 | – | – | – |
| [47] | C5.0 rules | 95.5 | – | – | – |
| [47] | SLIPPER (rules + AdaBoost) | 94.1 | – | – | – |
| [48] | kNN Global Alignment k=5 | 93.90 | – | – | – |
| [48] | SVM | 97 | – | – | – |
| [49] | C4.5 + Boosting (rule-based) | 94.7 | 96.46 | 92.81 | 94.84 |
| [50] | Naive Bayes | 94.80 | – | – | – |
| [50] | General Bayesian network (K2) | 96.22 | – | – | – |
| [51] | Convolutional NN | 96.18 | – | – | – |
| [30] | HDNA (Encoder II) | 93.4 | 96.7 | 91.5 | 92.15 |
| Our | Lev kNN k=27 | 84.82 | 88.31 | 96.75 | 77.64 |
| Our | Sym kNN R=1 k=425 | 96.71 | 94.16 | 98.70 | 96.98 |
| Our | Sym kNN R=1 s=1 k=170 | 90.45 | 83.77 | 94.81 | 91.54 |
| Our | Sym kNN R=2 k=375 | 91.24 | 87.01 | 97.40 | 90.33 |
| Our | HV+SVM R=1 BC=KS=100.0 opt | 97.03 | 98.05 | 97.40 | 96.37 |
| Our | HV+SVM R=1 m=11 std=0.289 | 95.63 | 93.83 | 97.00 | 95.82 |
| Our | HV+SVM R=1 m=111 std=0.326 | 95.85 | 93.32 | 97.01 | 96.48 |
| Our | HV+Prototypes R=1 m=111 std=0.483 | 94.09 | 96.27 | 98.43 | 91.06 |
| Our | HV+kNN R=1 k=425 m=11 std=0.333 | 96.16 | 95.22 | 98.56 | 95.48 |

As in [42]–[45], the measure Top-$n$ = $t_n/t$ was is used as an indicator of quality or accuracy, where $t_n$ is the number of cases where correct words are contained among the $n$ words of the dictionary most similar to the query, $t$ is the number of queries (i.e., the size of the test set). Two datasets were used: aspell[1] and wikipedia[2]. The tests contain misspellings for some English words and their correct spelling. Our results are obtained with the *corncob*[3] dictionary containing 58109 lowercase English words.

Fig. 3 shows the aspell Top-$n$ vs $R$ for $n$ = {1,10} and their average (Top-mean) for $n$ = 1...10. Here and thereafter, the dimension of HVs is $D$ = 10000. $R$ = 1 corresponds to the lack of similarity between letter HVs at adjacent positions. As $R$ increases, the results improve upto $R$ = 6–8, then deteriorate slowly.

Our results obtained using $\mathrm{sim}_{\mathrm{HV}}$ and $\mathrm{sim}_{\mathrm{Sym}}$ for various parameters are shown in Table I. For HVs, means and stds are given (over 50 realizations). The results of Word®, Ispell [42], Aspell [42], and the spellcheckers from [43]–[45] are also shown. All these spellcheckers work with single words, i.e., do not take into account the adjacent words. However, the results of [42]–[44] were obtained using methods specialized for English (using rules, word frequencies, etc.). On the contrary, our approach will naturally extend to other languages.

Only [45] worked with HV representations and *corncob*. However, they used the HVs of all 2-grams in the forward direction and in the backward direction, as well as all subsequences (i.e., non-adjacent letters) of two letters in the backward direction. This is in striking contrast to our HV representations, which reflect only the similarity of the same individual letters at nearby positions. Our results are at the level of [45]. Our best results were obtained for $s$ = 0 (no string shifts, see section IV-B-1). Increasing $s$ did not lead to a noticeable result change. Note that the similarity search using $\mathrm{dist}_{\mathrm{Lev}}$ and $\mathrm{dist}_{\mathrm{Lev}}$/max (divided by the length of the longer word) produced results that are inferior to ours.

## B. *Classification of molecular biology data*

Experiments were carried out on two Molecular Biology datasets from [46]. For hypervectors, we used the following classifiers: nearest neighbors *k*NN (mainly with $\mathrm{sim}_{\mathrm{HV,cos}}$), Prototypes, and linear SVM. In Prototypes, class prototypes were obtained by summing the HVs of all training samples from the class; their max similarity with the test HV was used. For SVM, in some cases, the parameters for a single realization of hypervectors were selected by optimization on the training set. The same parameters (or default) were used for multiple HV realizations.

*1) Splice junction recognition:* The Splice-junction Gene Sequences dataset [46] contains gene sequences for which one needs to recognize the class of splice junctions they correspond to: exon-intron (EI), intron-exon (IE), and no splice (Neither). Each sequence contains 60 nucleotides. The database consists of 3190 samples; 80% of each class was used for training and 20% for testing. Recognition results (accuracy) are shown in Table II. The results obtained using hypervectors are on a par with the results of other methods from [46]–[51]. Note that a direct comparison of the results is incorrect due to different data partitioning into training and test sets.

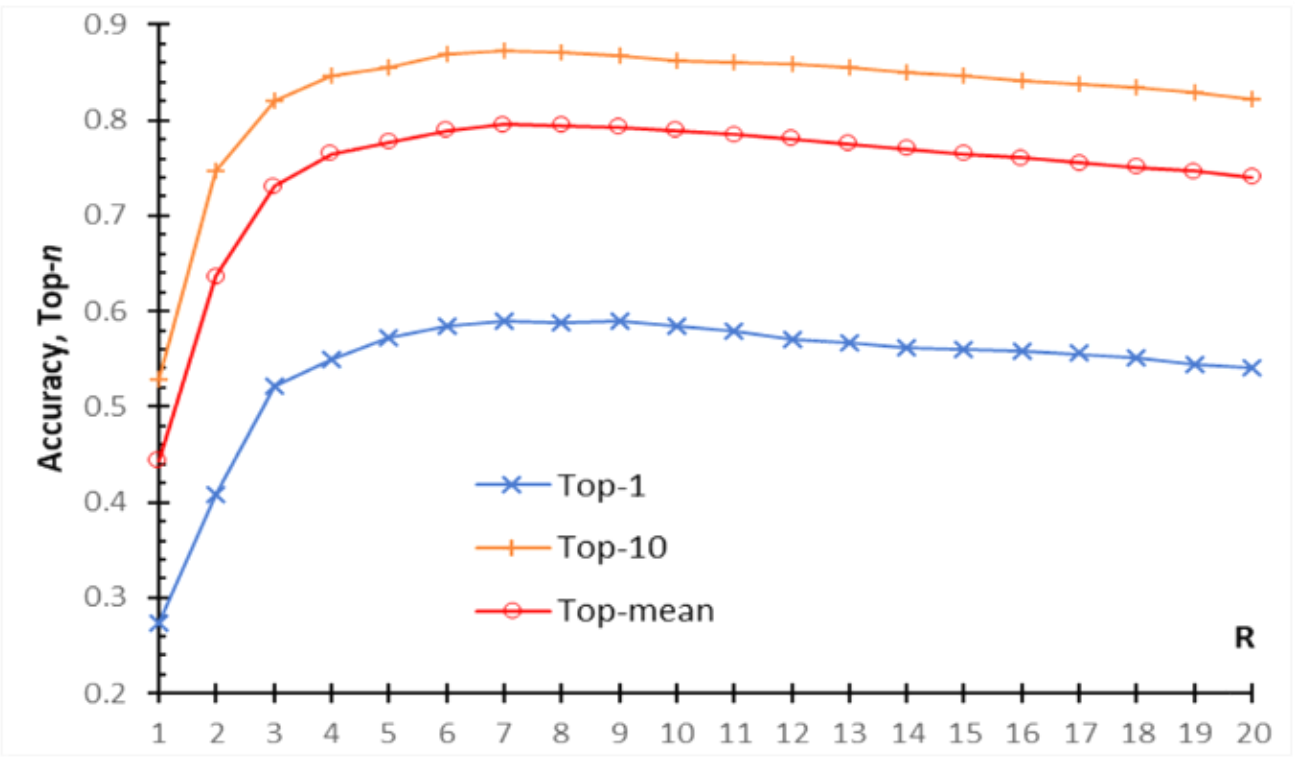

[1] http://aspell.net/test/cur/batch0.tab
[2] https://www.dcs.bbk.ac.uk/~ROGER/wikipedia.dat
[3] https://github.com/sibosop/speclib/blob/master/corncob_lowercase.txt

The best results were obtained for $R = 1$, $s = 0$. This corresponds to an element-wise comparison of the sequences. We explain this by the fact that the sequences in the database are well-aligned and the recognition result in this problem depends on the presence of certain nucleotides at strictly defined positions. Nevertheless, the introduced hypervector representations and similarity measures demonstrate competitive results for the selected parameters.

*2) Protein secondary structure prediction:* The Protein Secondary Structure dataset [46] contains some globular proteins data from [52], and the task is to predict their secondary structure: random-coil, beta-sheet, or alpha-helix. As input data, a window of 13 consecutive amino acids was used, which was shifted over proteins. For each window position and for the amino acid in the middle of the window, the task was to predict what secondary structure it is a part of within the protein. The training/test sets contained 18105/3520 samples. The prediction results are shown in Table III. The results of HV and linear SVM for $R = 1$ are at the level of 62.7% [52] obtained by multilayer perceptron for the same experimental design. Using $R = 2$ slightly improved the results obtained with $R = 1$.

Note, that the results obtained on this dataset under this very setup using other methods are inferior to ours (see, e.g., [52]). To improve the results in this and similar tasks, some techniques after [52] used additional information, such as the "similarity" of amino acids, etc. This information can be taken into account in the varied similarity of HVs representing different amino acids, however, this is beyond the scope of this paper (see also Discussion).

The results of this section show that not all string processing tasks benefit from accounting for symbol insertions/deletions (in our approach, regulated by $R$) and string shifts (regulated by $s$). E.g., for the Splice-junction dataset, $R > 1$ worsened the results, and for the Secondary-structure dataset, $R = 2$ only slightly improved them. However, we see that the HV representations of the vector data with symbolic (nominal) components provide worthy classification results using linear vector classifiers.

The tasks in which the results depend significantly on both parameters $R$ and $s$ are considered in the next section.

TABLE III
PREDICTION ACCURACY
ON THE PROTEIN SECONDARY STRUCTURE DATASET

| Method | Total % | Coil % | Sheet % | Helix % |
|---|---|---|---|---|
| Backprop 13 amino acids input [52] | 62.7 | – | – | – |
| Lev k=37 (of 100) | 58.72 | 95.73 | 21.32 | 5.614 |
| HV+Prototypes m=11 R=1 | 54.23 | 50.24 | 62.02 | 55.66 |
| HV+SVM R = 1 m=1 opt | 62.78 | 83.05 | 45.70 | 30.08 |
| HV+SVM R = 2 m=1 opt | 63.21 | 83.05 | 46.29 | 31.41 |
| HV+SVM R = 3 m=1 opt | 62.47 | 82.89 | 45.23 | 29.55 |
| HV+SVM, R =1 m =11 | 62.67 | 84.53 | 44.12 | 27.53 |
| std | 0.0934 | 0.142 | 0.265 | 0.223 |
| HV+SVM, R =2 m =11 | 62.86 | 82.97 | 45.65 | 30.69 |
| std | 0.0567 | 0.0781 | 0.1432 | 0.1457 |
| HV+kNN k=28 m=111 R=1 | 57.39 | 90.68 | 22.32 | 11.64 |
| std | 0.5261 | 0.5101 | 1.550 | 1.011 |
| HV+kNN k=28 m=111 R=2 | 58.36 | 88.49 | 27.68 | 15.73 |
| std | 0.3682 | 0.3277 | 1.036 | 0.8358 |

### C. *Modeling visual string identification by humans*

Here we present the results of experiments on the similarity of words using their hypervector representation. The results are compared to those obtained by psycholinguists for human subjects and provided in [53],[54]. Those experiments investigated priming for visual (printed) words in humans.

*1) Modeling restrictions on the perception of word similarity:* In [29], the properties of visual word similarity obtained by psycholinguists in experiments with human subjects have been summarized and classified into 4 types of constraints, i.e., stability (similarity of a string to any other is less than to itself); edge effect (the greater importance of the outer letters coincidence vs the inner ones); transposed letter (TL) effects (transposing letters reduces similarity less than replacing them with others); relative position (RP) (breaking the absolute letter order while keeping the relative one still gives effective priming).

Table IV shows which constraints on human perception of visual word similarity are satisfied in various models. Our results were obtained for $\text{sim}_{\text{Sym}}$ and $\text{sim}_{\text{HV}}$ ($D = 10000$, $m = 11$, 50 realizations). To reflect the edge effect, we used the "db" option: the HVs were formed in a special way equivalent to the HV representation of strings with doubled first and last letters.

For $s > 2$, the results coincided with $s = 2$. The best fit to the human constraints was for $R = \{2,3\}$, $s = 2$. For $\text{sim}_{\text{Sym,cos}}$ and for $\text{sim}_{\text{HV,Simp}}$ all constraints are satisfied for $R = 3$, $s = 2$.

For comparison, the results of other models are shown:

TABLE IV
CONSTRAINTS ON HUMAN PERCEPTION OF VISUAL WORD SIMILARITY
THAT ARE SATISFIED BY VARIOUS MODELS

| Study | Method | Stabi-lity | Edge Eff. | Loc. TL | Glob. TL | Dist. TL | Comp. TL | Dist. RP | Rep. RP |
|---|---|---|---|---|---|---|---|---|---|
| [29] | HV BSC Slot coding | ✓ | – | ✓ | ✓ | – | ✓ | – | – |
| [29] | HV BSC COB | ✓ | – | ✓ | – | – | – | ✓ | ✓ |
| [29] | HV BSC UOB | ✓ | ✓ | ✓ | – | – | – | ✓ | ✓ |
| [29] | HV BSC LCD | ✓ | ✓ | ✓ | – | – | – | ✓ | – |
| [29] | Spatial Coding | – | ✓ | ✓ | ✓ | ✓ | ✓ | ✓ | ✓ |
| [29] | Seriol | ✓ | – | ✓ | – | ✓ | ✓ | ✓ | – |
| [32] | HV BSC bin 2X | ✓ | – | ✓ | – | – | ✓ | ✓ | ✓ |
| [32] | HV HRR real 2X | ✓ | – | ✓ | – | – | ✓ | ✓ | ✓ |
| [32] | 1–Lev/AddedLength | ✓ | – | ✓ | ✓ | – | – | ✓ | ✓ |
| [54] | HV HRR Terminal | ✓ | ✓ | ✓ | ✓ | ✓ | ✓ | ✓ | ✓ |
| Our | minus Lev | ✓ | – | – | ✓ | – | ✓ | ✓ | ✓ |
| Our | Sym cos R=2 s=2 | ✓ | – | ✓ | ✓ | – | ✓ | ✓ | ✓ |
| Our | Sym cos R=3 s=2 | ✓ | – | ✓ | – | ✓ | ✓ | ✓ | ✓ |
| Our | Sym cos R=2 s=2 db | ✓ | ✓ | ✓ | ✓ | – | ✓ | ✓ | ✓ |
| Our | Sym cos R=3 s=2 db | ✓ | ✓ | ✓ | ✓ | ✓ | ✓ | ✓ | ✓ |
| Our | HV cos R=2 s=2 db | ✓ | ✓ | ✓ | ✓ | – | ✓ | ✓ | ✓ |
| Our | HV cos R=3 s=2 db | ✓ | ✓ | ✓ | ✓ | ✓ | ✓ | ✓ | ? |
| Our | HV Simp R=3 s=2 db | ✓ | ✓ | ✓ | ✓ | ✓ | ✓ | ✓ | ✓ |

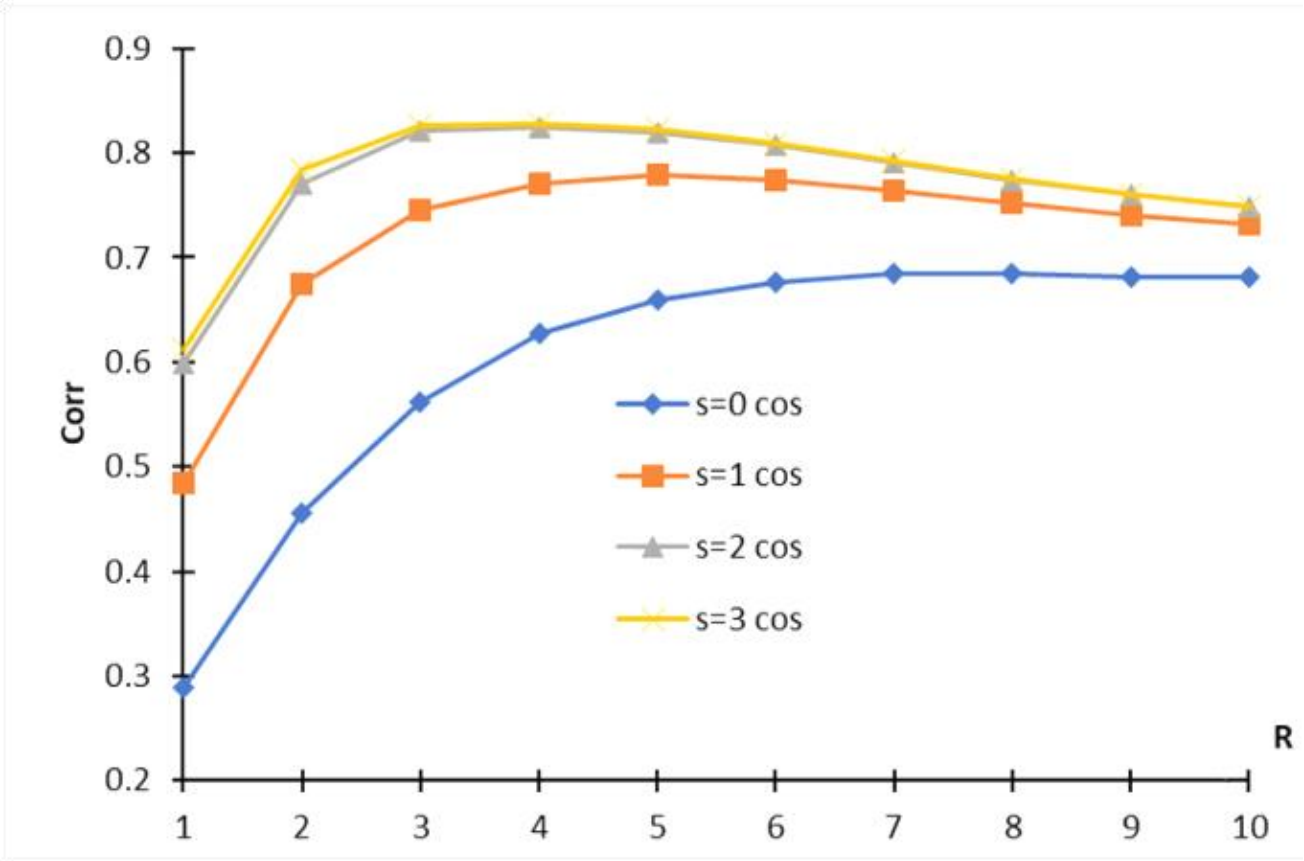

Fig. 4. Mean values of the Pearson correlation coefficient Corr between hypervector similarity values and the times of the forward priming vs *R*.

Hannagan et al. [29] used the HV representations of the BSC model [20]. Slot: superposition of HVs obtained by binding HVs of each letter and its (random) position HV. COB: all subsequences of two letters with a position difference of up to 3. UOB: all subsequences of two letters. LCD: a combination of Slot and COB. [29] also tested non-hypervector models: Seriol and Spatial [53].

Cohen et al. [32] used the BSC model and real- and complex-valued HVs of HRR [19]. Position HVs were correlated, and their similarity decreased linearly along the length of a word.

Cox et al. [54] proposed the TR string representation scheme. It used the representation of letters and 2-grams without position, as well as the representation of the positions of letters and 2-grams relative to the terminal letters of the word. The TR scheme was implemented in the HRR model. It met all the constraints from [29].

*2) Modeling restrictions on the perception of word similarity:* In [55], the experimental data on the visual word identification by humans were adapted from [53]. 45 pairs of prime-target strings were obtained, for which there exist the times of human word identification under different types of priming.

Fig. 4 shows the average value of the Pearson correlation coefficient Corr (between the $\mathrm{sim}_{\mathrm{HV}}$ values and priming times) vs $R$ for different $s$ ($D = 10000$, $m = 111$, 50 HV realizations). It can be seen that the value of Corr depends substantially on $R$ and $s$, the maximum values were obtained at $R = 3$ and $s = 2$.

Table V shows the Corr between the times of human identification and the values of similarity. The results for HVs were obtained for $R = 3$, $s = 2$ ($D = 10000$, $m = 111$, 50 realizations) and for similarity measures $\mathrm{sim}_{\mathrm{HV,Jac}}$, $\mathrm{sim}_{\mathrm{HV,cos}}$, $\mathrm{sim}_{\mathrm{HV,Simp}}$. The db option was used. We also provide results for $\mathrm{dist}_{\mathrm{Lev}}$ and $\mathrm{sim}_{\mathrm{Sym,Simp}}$ ($R = 3$, $s = 2$) without the db option.

TABLE V
THE PEARSON CORRELATION COEFFICIENT CORR BETWEEN 45 WORD PAIRS SIMILARITY VALUES AND THE TIMES OF THE FORWARD PRIMING

| Method | Spatial Coding | GvH UOB | Kernel UOB | 3Wild Card | Lev/ max | Sym Simp | HV Jac | HV cos | HV Simp |
|---|---|---|---|---|---|---|---|---|---|
| Corr | 0.732 | 0.673 | 0.747 | 0.797 | 0.834 | 0.843 | 0.831 | 0.822 | 0.866 |

The results from [55], are shown as well, where a vector is formed for a string with the components corresponding to certain combinations of its letters. Spatial Coding: adapted from [53]. GvH UOB: All subsequences of two letters are used. Kernel UOB (Gappy String kernel): uses counters of all subsequences of two letters within a window. 3-WildCard (gappy kernel): kernel string similarity [55] (all subsequences of two letters are padded with * in all acceptable positions, the vector contains the frequency of each obtained combination of three symbols).

It can be seen that with the proper parameters, the results of hypervector similarity measures are competitive with other best results, such as $\mathrm{dist}_{\mathrm{Lev}}$ and $\mathrm{sim}_{\mathrm{Sym}}$.

## VI. DISCUSSION

The paper proposes a hypervector representation of sequences that is equivariant with respect to sequence shifts and preserves the similarity of identical sequence elements at nearby positions. The case of symbol strings is considered in detail. The hypervector representation of a string is formed from the hypervectors of its symbols at their positions. A similarity measure of symbol strings is also proposed that does not use hypervectors but approximates their similarity.

The proposed methods were explored in diverse tasks where strings are used: spellchecking, classification of molecular biology data, modeling of human perception of word similarity. The results obtained are on a par with the results of other methods that, however, additionally use *n*-gram or subsequence representations of strings or some other domain knowledge.

*1) Other types of sequence elements:* Our approach allows using various types of sequence elements, i.e., the data types for which hypervector representations are known can be used. They include numeric scalars or vectors, *n*-grams, other sequences, graphs, etc. Moreover, the proposed methods do not demand sequence elements to be in contiguous positions, as in strings. These modifications may require some method adaptations, such as increasing hypervector dimensionality or/and adjusting parameters and techniques.

Also, our approach can be applied to representing vectors with components that are integers in a fixed range (symbols would correspond to the components of the vector, whereas positions would correspond to the components' values).

*2) Equivariance:* The equivariance of representations is a desirable property, at least for the following reasons:

- In the equivariant representations, the information about the transformation $S(x)$ for which the representation is obtained is preserved and available for further processing. For example, a hypervector representation equivariant with respect to a sequence shift preserves information about the position of the sequence. This contrasts with the invariant representation, where such information is lost.
- Ensuring equivariance in hypervector representation opens up the possibility to perform their further equivariance-preserving transformations.
- From an equivariant representation, an invariant one can be obtained. E.g., this could be done by superposition of the hypervectors obtained for all the transformations with respect to which invariance is required.

• The system gets the ability to operate with the internal representations of objects without using the objects themselves.

• Obtaining the hypervector of the transformed object as $T(F(x))$ is computationally more efficient than as $F(S(x))$, if $T$ is easier to calculate than $F$ and there exists previously obtained object hypervector $F(x)$.

• The absence of computing and energy costs for the execution of $F(S(x))$ is important in case of limited resources, e.g., in edge computing.

We also foresee other interesting effects from equivariant hypervector representations, both in line with DNNs [26] and beyond.

*3) Directions for future research:* In this paper, the proposed approach for hypervector representation of sequences has been detailed and tested for the case of rather short symbolic strings and for the HDC/VSA model of Sparse Binary Distributed Representations [14], [15], [18]. Areas for further research include the following extensions:

• other HDC/VSA models;
• long sequences; hierarchical sequences;
• other data types (besides sequences);
• other types of equivariance (besides shifts);
• other types of application tasks;
• interplay with DNNs.

Some of these extensions look rather straightforward, some will probably require more research and novel solutions.

Concerning further progress in the HDC/VSA field, one promising direction is representing different types of data in a single hypervector [18], [8]. For example, different descriptors for a single image [8] or different modalities of object representation [18]. When using permutative hypervector representations, this would require applying different permutations. Unlike the formation of distributed vector representations in DNNs, no training is needed to form such hypervector representations in HDC/VSA.

## Acknowledgment

To be added